%% file: ms.tex
\documentclass[10pt,twocolumn,letterpaper]{article}

\usepackage{cvpr}
\usepackage{times}
\usepackage{epsfig}
\usepackage{graphicx}
\usepackage{amsmath}
\usepackage{amssymb}
\usepackage{multirow}
\usepackage{booktabs}
\usepackage{caption}
\usepackage{subcaption}
\usepackage{enumitem}

\usepackage[pagebackref=true,breaklinks=true,letterpaper=true,colorlinks,bookmarks=false]{hyperref}

\cvprfinalcopy 

\def\cvprPaperID{4722} 

\ifcvprfinal\fi
\begin{document}

\title{Effective Aesthetics Prediction with Multi-level Spatially Pooled Features}



\author{Vlad Hosu \hspace{1.9cm} Bastian Goldl\"{u}cke \hspace{1.9cm} Dietmar Saupe\\
University of Konstanz, Germany\\
{\tt\small \{vlad.hosu, bastian.goldluecke, dietmar.saupe\}@uni-konstanz.de}\\
}

\maketitle


\begin{abstract}
We propose an effective deep learning approach to aesthetics quality assessment that relies on a new type of pre-trained features, and apply it to the AVA data set, the currently largest aesthetics database. While all previous approaches miss some of the information in the original images, due to taking small crops, down-scaling or warping the originals during training, we propose the first method that efficiently supports full resolution images as an input, and can be trained on variable input sizes. This allows us to significantly improve upon the state of the art, increasing the Spearman rank-order correlation coefficient (SRCC) of ground-truth mean opinion scores (MOS) from the existing best reported of 0.612 to 0.756. To achieve this performance, we extract multi-level spatially pooled (MLSP) features from all convolutional blocks of a pre-trained InceptionResNet-v2 network, and train a custom shallow Convolutional Neural Network (CNN) architecture on these new features.

\end{abstract}

\section{Introduction}


Aesthetics quality assessment~(AQA) is an interesting, yet very difficult task to perform algorithmically. It involves predicting subjective opinions, i.e., aesthetics ratings, or distributions thereof. The complexity of the task is due to the many influencing factors for human judgments. An image's aesthetic quality, in addition to photography rules-of-thumb, is influenced by affective and personal preferences, for example for different content types or style. Earlier studies~\cite{datta2006studying, ke2006design} have tried to encode photography rules as features that were used for training AQA models. Later works use different hand-crafted features~\cite{luo2008photo, tang2013content, dhar2011high}  or more general purpose ones~\cite{marchesotti2011assessing, su2011scenic}.

A limitation of these traditional machine learning approaches is the representational power of the extracted features, and they are now outperformed by deep learning, which has been shown to be very suitable for AQA~\cite{lu_deep_2015,lu2015rating,mai2016composition,kong2016photo,ma_alamp_2017,hii_multigap_2017,talebi_nima_2017,schwarz2018will,karayev2013recognizing}. The de facto standard for training and evaluation is the large-scale database for Aesthetic Visual Analysis (AVA) \cite{murray_ava_2012}. It consists of about 250 thousand images, collected from a photography community\footnote{dpchallenge.com}, each rated by an average of 210 users. 

Unfortunately, approaches based on deep learning typically need to perform a combination of rescaling and/or cropping to fit to the input shape of the chosen Deep Neural Network (DNN).
The reason is that for efficient computation, GPUs need to process fixed sized blocks of data at a time, such as batches of images at the same resolution. 
%
However, AVA contains thousands of unique resolutions. Even for DNN architectures which can accept arbitrary input image resolutions, training would be extremely inefficient due to the small batch sizes required. Not only would processing be much slower, but learning performance would also suffer for models that employ batch normalization, such as Inception-v3.
Moreover, the AVA database contains high resolution images, with an average width $\times$ height of $629\times497$ and maximum of $800\times800$ pixels. Training DNN models at this size is very resource intensive, requiring images in the database to be resampled.
An obvious solution is to rescale all images to the DNN input resolution when training, which either removes high-frequency information when down-scaling, or introduces artificial blurs when up-scaling. Thus, this approach removes or masks some information that the models could learn from, leading to a lower performance~\cite{mai2016composition}.
Even if we proportionally down-scale large images to at least keep their aspect ratio, some information is still lost, and it would not alleviate the problem of different input resolution.
Consequently, several existing works~\cite{hii_multigap_2017, lu_rapid_2014} have already noted the drawbacks of learning to predict the aesthetics scores from images that are not of the same resolution as the original. Images that are downsized, stretched, or cropped do not contain the same information as the higher resolution image that was originally assessed by human observers. Thus, predictive performance is negatively affected.

On the other hand, previous works~\cite{zhang_unreasonable_2018, gao_blind_2018} have shown the effectiveness of  multi-level pre-trained features (models trained on ImageNet) to predict perceptual judgments, either for image similarity or image quality assessment (IQA). These have employed earlier DNN models such as VGG16. In \cite{hii_multigap_2017},  multi-level features were extracted from the more modern Inception architecture. The authors did not use pre-trained features directly, but fine-tuned the network and then used the updated features.
%

\input{fig_overview}

%
\textbf{Contributions.}
%
%
%
In this work, we devise an efficient staged training approach for AQA relying on a new type of perceptual features (MLSP). 
We show that we can efficiently learn from these very high-dimensional features ($5\times 5\times 16,928$), far surpassing the state of the art with respect to correlation metrics. To achieve this, we extract and store fixed sized features from variable resolution images with augmentation, Fig.~\ref{fig:overview}. We propose multiple head architectures that give better results either for narrow ($1\times 1\times b$) or wide ($5\times 5\times b$) MLSP features.

Variable, large resolution inputs have been the main stumbling block of all existing approaches on AVA, limiting training to low resolution images or small batch sizes. Compared to end-to-end training, staged training significantly reduces GPU memory requirements, standardizes variable resolution inputs enabling larger batch sizes, which leads to much faster training speeds (20x--200x speedup). This has several benefits:

\begin{itemize}[leftmargin=*]
    \item \textbf{Effectiveness}: we significantly improve upon the state-of-the-art in AQA on AVA, particularly for correlation metrics: previous 0.612 SRCC \cite{talebi_nima_2017} vs ours 0.756 SRCC.
    \item \textbf{Reduced GPU memory}: MLSP features are compact, and independent of the input image resolution. The head networks required by our approach are shallow making training much more memory-efficient.
    \item \textbf{Speed}: training for an epoch on AVA for wide MLSP features takes about 10 minutes, while for narrow MLSP features it is less than a minute. Fine-tuning on average sized images ($629\times497$ pixels) takes about 4.5 hours on one GPU. 
\end{itemize}

%
%
\section{Related work}

State-of-the-art Convolutional Neural Network (CNN) architectures for image classification are designed for small images below $300\times300$ pixels, all at the same resolution.
Thus, deep learning for AQA has faced two major challenges. First, the AVA database contains over 25 thousand unique image resolutions, second, images are large, up to $800\times800$ pixels.
Since users rate the original images, any transformation applied to an original
%
%
does not preserve the available information. Many existing DNN AQA methods have tried to overcome these two main limitations by designing multi-column architectures that take several small resolution patches or fixed size rescaled images (usually $224\times224$) as inputs and aggregate their contributions to the aesthetics score.

Lu et al.~\cite{lu2015rating} employ a double-column DNN architecture to support the integration of both a global rescaled view, $224\times224$ randomly cropped from the $256\times256$ downsized original, and a local crop of $224\times224$ extracted from the original. In~\cite{lu_deep_2015}, they extend the initial work to use multiple columns that take random $256\times256$ patches from the original image. Both networks employ custom shallow shared-weights CNNs for each column. Ma et al.\ \cite{ma_alamp_2017} further extend Lu et al.'s work \cite{lu_deep_2015} by selecting $224\times224$ patches non-randomly based on image saliency and aggregating them while modeling their relative layout. Instead of taking patches, Mai et al. \cite{mai2016composition} use a multi-column architecture to integrate five spatial pooling sizes, which allows for testing with variable resolution images. Their base column architecture is VGG-16. Our main improvement over these approaches is to significantly simplify the architecture and bypass the requirement for taking small patches altogether. We train on features extracted directly from the original images, alleviating the limitations of previous works.

Talebi et al. \cite{talebi_nima_2017} have shown that simple extensions to existing image classification architectures can work surprisingly well. They rely on VGG16, MobileNet, and Inception-v2 base architectures, and replace the classification with a fully-connected regression head that predicts the distributions of ratings for each image. While training on $224\times224$ crops from $256\times256$ rescaled images in AVA, they showed promising results. We take the same approach to augmentation when fine-tuning base networks with our proposed regression head. However, when training on features from the original images, we extract crops that are proportional to each image, at 87.5\% the width and height.

Another less strongly related branch of methods relies on ranking-type losses to encode the relationships between the aesthetic quality of images. Kong et al.\ \cite{kong2016photo} propose to rank the aesthetics of photos by training on AVA with a rank-based loss function. They introduced a two column architecture (AlexNet-type base CNN), learning the difference of the aesthetic scores between two input images. They trained on low resolution $227\times227$ random crops from $256\times256$ rescaled originals. Schwarz et al.\ \cite{schwarz2018will} use a triplet loss, with a ResNet base architecture taking inputs at $224\times224$ pixels.
Kong et al.\ \cite{kong2016photo} are the first to report the SRCC metric on AVA, which is a natural way to evaluate ranking loss. Both works suggest that optimizing for ranking is important.

Aesthetic quality involves both low level factors such as image degradations (blur, sharpness, noise, artifacts) as well as higher level content-related factors. The authors of~\cite{hii_multigap_2017, karayev2013recognizing} therefore propose to employ concatenated features from multiple levels of a network. 
Karayev et al.\ \cite{karayev2013recognizing} compute content features from the last two levels of a DeCAF network (DeCAF5 and DeCAF6) and have shown that these features extracted from a pre-trained network on ImageNet perform well when applied to recognizing image style.
Hii et al. \cite{hii_multigap_2017} concatenate global average pooled (GAP) activations from the last 5 blocks of an Inception \cite{szegedy2015going} network and fine-tune the architecture for binary classification. Both of these approaches suggest that multi-level features can be helpful in predicting aesthetics. However, in contrast to our approach, they only consider some of the latter levels in the network, and extract a small set of globally pooled features.

In addition to these aesthetics-related works, others have considered using multi-level features for perceptual similarity \cite{zhang_unreasonable_2018} or image quality assessment \cite{gao_blind_2018}. They use ImageNet pre-trained features directly and a form of global pooling by means of one or more statistics, e.g., mean, max, min. In our work, we study spatially pooled features from multiple levels in Inception-v3 and InceptionResNet-v2 DNNs in-depth, and investigate how to best use them for assessing aesthetics.

%
%
\section{Network architecture and training}

\subsection{Feature extraction}

Modern deep neural networks for object classification are structured in multiple sequential operation blocks that each have a similar structure.
For inception networks~\cite{szegedy2015going, szegedy2017inception}, these are often called inception modules. We study the suitability of the Inception-v3 and InceptionResNet-v2 blocks for AQA more in-depth by extracting both types of proposed MLSP features.

\begin{figure}[!b]
    \centering
    \includegraphics[width=\linewidth]{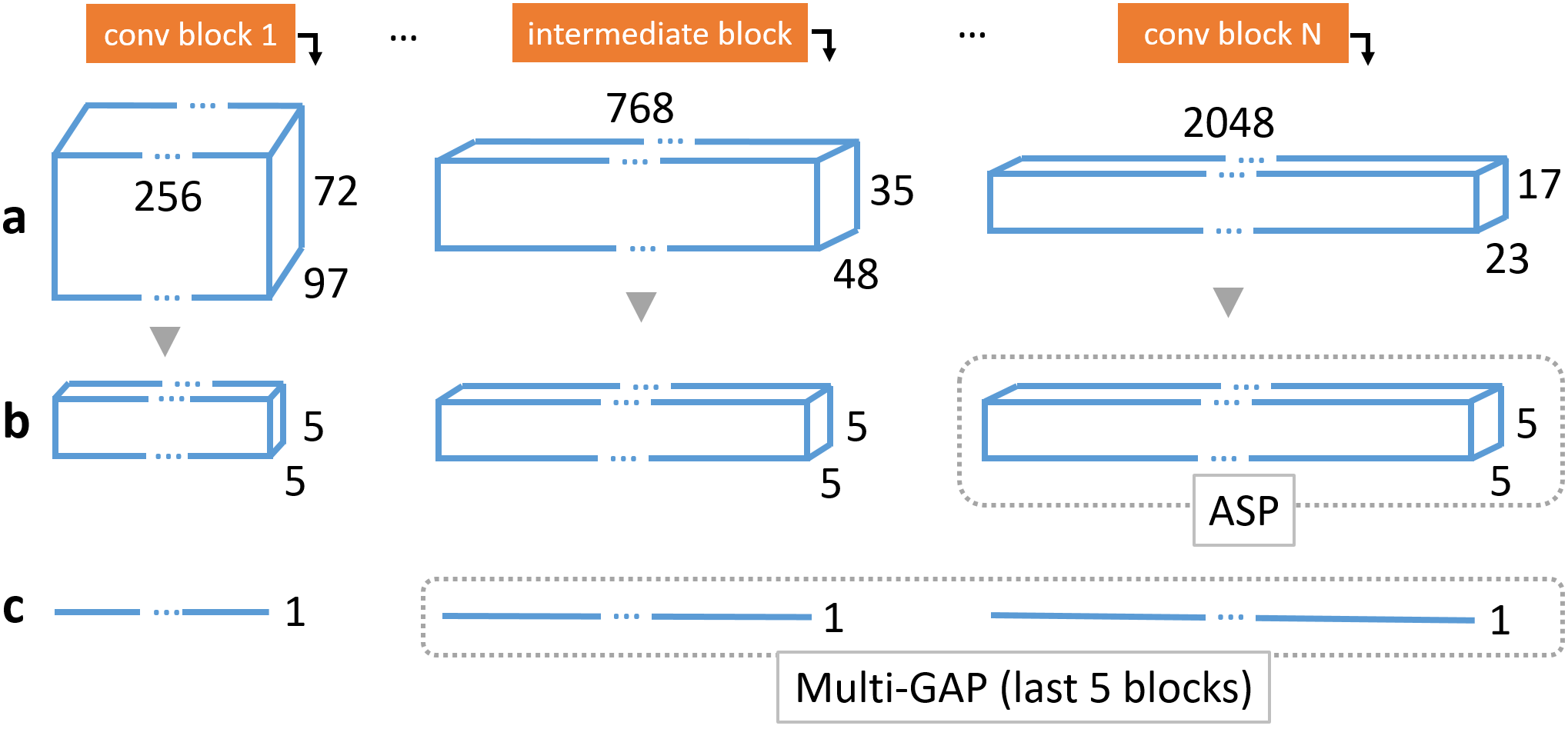}
    \caption{\textbf{a)} 3 of 11 activation blocks from an Inception-v3 CNN with an input image size of $800\times600$ pixels. \textbf{b)}, \textbf{c)} MLSP features, inspired by ASP \cite{mai2016composition} (originally used VGG-16), and Multi-GAP \cite{hii_multigap_2017}.}
    \vspace{-10pt}
    \label{fig:mlsp-sota}
\end{figure}

In Fig.\ \ref{fig:overview}, we show a general description of the approach for an Inception-type network. The pooling operation is slightly different between the two feature types: one resizes each activation block output of an inception module to a fixed spatial resolution: $1\times1$ global average pooling (GAP) for narrow features, and $5\times5$ spatial average pooling for wide MLSP features, Fig. \ref{fig:mlsp-sota}. %
We compute the spatial pooling by resizing feature blocks using area interpolation (the same as INTER\_AREA in OpenCV). The resized features are concatenated along their kernel axis. There are 11 blocks in Inception-v3 (10,048 kernels), and 43 in InceptionResNet-v2 (16,928 kernels).

Multiple augmentations are considered for each image in the AVA data set, and their corresponding features are stored. We augment images with proportional crops at 87.5\% of width and height, e.g., $224\times224$ from $256\times256$, covering the four corners of each image, together with a mirroring augmentation (on and off). The 8 sets of features are stored in a single HDF5 file for all images.
In our tests there was no significant decrease in performance (train and test) when converting features from 32 to 16 bit floating point numbers, so we use the smaller type to save memory.
During training AQA on features, a single random augmentation per epoch is drawn for each image from among the 8 stored (4 crops, 2 flips). At test time, the final score for an image is calculated by averaging the predictions over all stored augmentations. This is different to when we do fine-tuning of the base network, where we use crops at completely random locations instead and random flips. At test time in this latter situation, we average over 20 random augmentations obtained in this way. We noticed there is generally no improvement in performance if we use more than 20 augmentations.

\input{fig_3fc_head_arch.tex}
\subsection{AQA from deep image features}

In this section, we design different architectures for learning AQA from deep features. 
In principle, they can be stacked on top of the base network for feature extraction and trained end-to-end. However, our experiments at low image resolution show that there is no significant performance gain in doing so, see Table~\ref{fig:performance-of-blocks} later on, while at high image resolutions this approach would be too demanding on GPU memory.
It turns out that narrow MLSP features do well with simpler fully-connected architectures, whereas we need another approach to handle the spatial component of the wide MLSP features.

A common component of many of our proposed networks is a custom fully-connected head (3FC) that is tied directly to predicting the mean opinion score (MOS) for each image. Throughout the work we use a Mean Squared Error (MSE) loss in our architectures. The 3FC head, Fig.~\ref{fig:3fc_head_arch}, uses dropout to reduce over-fitting and batch normalization to improve the smoothness of the convergence profile, i.e., the loss curve.
It is assigned a special symbol, see Fig.~\ref{fig:3fc_head_arch}, which is used in the following figures.

\input{fig_narrow_head_arch.tex}
\subsubsection{Narrow MLSP feature architectures}

We present three architectures that we designed to work best as regression heads on top of narrow MLSP features:

\textbf{Single-1FC}: The most simple network is similar to what Hii et al. \cite{hii_multigap_2017} introduced: adding a single fully-connected layer on top of the concatenated Global Average Pooling (GAP). We modify it by using a single score (MOS) predictor, see Fig.~\ref{fig:narrow_head_arch}(a). This model performs fairly well given its simplicity, but it can be easily improved.

\textbf{Single-3FC}: Here, we incorporate the 3FC component in the head architecture, as shown in Fig.~\ref{fig:narrow_head_arch}(b). This approach results in a further performance improvement.

\textbf{Multi-3FC}: Inspired by the BLINDER approach \cite{gao_blind_2018}, where the authors train independent Support Vector Regression models on top of each block of pooled features from all levels of a VGG network, we propose a fully-connected architecture that first learns from individual feature blocks and then linearly combines the results into the final prediction, see Fig.~\ref{fig:forest_arch}. Feature blocks are extracted from the output of each Inception module. This architecture, even though having more capacity and more granular information than Single-3FC, performs about the same.

We have considered other variations on the presented models, including changes in number of neurons, fully connected structure, use of batch normalization, dropout rates, etc. The reported parameters worked best for us.

\input{fig_forest_arch.tex}

\subsubsection{Wide MLSP feature architecture}

\textbf{Pool-3FC}: for wide MLSP features, we take a slightly different approach, as shown in Fig.~\ref{fig:spatial_head_architecture}. First we need to narrow down the features to a size that can be handled efficiently by fully connected components. We do this using a modified Inception module, where we only have three columns: $1\times1$, $3\times3$ convolutions, and average pooling with the corresponding $1\times1$ convolution to reduce the output size. After trying different numbers of kernels, we got the best results for 1024 per column.

\input{fig_spatial_head_architecture.tex}

\begin{table}[]
\centering
\resizebox{\linewidth}{!}{
\begin{tabular}{@{}llll@{}}
\toprule
Model name & Fig                                 & Summary                          \\ \midrule
Single-1FC & \ref{fig:narrow_head_arch} (a)        & 1 fully-connected (fc) layer     \\
Single-3FC & \ref{fig:narrow_head_arch} (b)        & 3 stacked fc + batch norm + dropout  \\
Multi-3FC  & \ref{fig:forest_arch}               & Single-3FC for each GAP block \\
Pool-3FC   & \ref{fig:spatial_head_architecture} & \begin{tabular}[c]{@{}l@{}}Inception-type module + Single-3FC\end{tabular}       \\ \bottomrule
\end{tabular}
}
\caption{Summary of proposed architectures. The Inception-type module is modified from Inception-v3. }
\label{tab:architectures}
\end{table}

\subsection{Training}

All our networks are trained on the AVA dataset \cite{murray_ava_2012}, and tested on the same random subset as previously used in the literature. 
This test set consists of 20,000 images, however we were only able to read correctly 19,928 of them. The remaining 235,574 images of AVA are further randomly split into a training (95\%) and validation set (5\%). The validation set is used to choose the best performing model, i.e. the one with minimum loss, via early stopping. The loss of the network is the MSE between the predicted and the ground-truth MOS for each image.

The initial learning rate used in training is $10^{-4}$, which is divided by 10 every 20 epochs, or when the loss does not decrease within 5 epochs, whichever comes first. The lowest learning rate we used was $10^{-6}$. At each change in learning rate the previous best model is loaded. The Adam optimizer \cite{kinga2015method} is used, with all hyper-parameters except learning rate set to their defaults. When learning AQA from our pre-trained features, we use a batch size of 128, and when fine-tuning models we use 32 images per batch. Experiments are performed on a single Nvidia Titan Xp GPU, using Keras with the Tensorflow backend to implement our models.

%
%
\section{Results}

\subsection{Rescaled input images}
In a first step, we only evaluate the performance of the different head architectures
from the previous section, and train on~$224\times224$ crops from images which have been rescaled to the fixed resolution of~$256\times256$.
The purpose is to compare several options for learning AQA from features, while having images which are small enough so it is still possible to fine-tune the entire architecture end-to-end without having to extract and store features as a first step.
Results and more details can be observed in Table~\ref{tab:feature-architectures}.
Notably, the optimal performance obtained with either end-to-end fine-tuning or learning just from pre-trained features is practically the same.
We only start to see a difference between these two if we use fewer features, for instance, Single-1FC (5) performs better when fine-tuned than when doing feature learning.
This indicates that learning from comprehensive features may generally be as good as fine-tuning, even when generalizing to larger resolution images. Note that images are augmented during fine-tuning by taking random crops and flips, whereas for learning from pre-trained features we only use four fixed position crops and flips, potentially putting the latter at a slight disadvantage. 
When testing on feature based models, we aggregate predictions over all stored augmentations, while for fine-tuned models we aggregate over 20 random augmentations.

%
The best model for learning from pre-trained features is Single-3FC, which is simpler than Multi-3FC. For this reason, later on, we only consider Single-3FC when comparing different approaches to feature extraction.

\begin{table}[h]
\centering
\resizebox{\linewidth}{!}{
\begin{tabular}{@{}lcccc@{}}
\toprule
Head architecture      & Finetune   & SRCC  & PLCC  & Accuracy  \\ \midrule
Single-3FC (-aug)      & no   & 0.668 & 0.676 & 79.2\%          \\
Single-3FC             & no   & 0.676 & 0.682 & 79.3\%          \\
Multi-3FC              & no   & 0.673 & 0.681 & 79.1\%          \\
Single-1FC (all)       & no   & 0.644 & 0.652 & 77.9\%          \\
Single-1FC (5)         & no   & 0.570 & 0.578 & 75.8\%          \\ \midrule
Single-3FC (-aug)      & yes  & 0.607 & 0.583 & 77.1\%          \\
Single-3FC             & yes  & 0.658 & 0.663 & 78.5\%          \\
Multi-3FC              & yes  & 0.655 & 0.660 & 78.4\%          \\
Single-3FC (+do)       & yes  & 0.675 & 0.682 & 79.1\%          \\
Single-1FC (all)       & yes  & 0.680 & 0.684 & 79.5\%          \\ 
Single-1FC (5)         & yes  & 0.675 & 0.681 & 79.1\%          \\\bottomrule
\end{tabular}
}
\caption{Performance of different architectures, trained on $224\times224$ crops from $256\times256$ rescaled originals, for direct learning from features (Finetune = no) or with fine-tuning. All are trained on narrow MLSP features (11 blocks) from Inception-v3, except for Single-1FC (5) which mimics the approach taken by multiGAP  \cite{hii_multigap_2017}, modified for single score prediction, using only the last 5 blocks. For Single-3FC (+do) the dropout rates are increased to (0.5, 0.5, 0.75). Single-3FC (-aug) does not use augmentation, but is simply trained on features from $256\times256$ rescaled originals.}
\vspace{-2mm}
\label{tab:feature-architectures}
\end{table}

\begin{figure}[h]
    \centering
    \includegraphics[width=\linewidth]{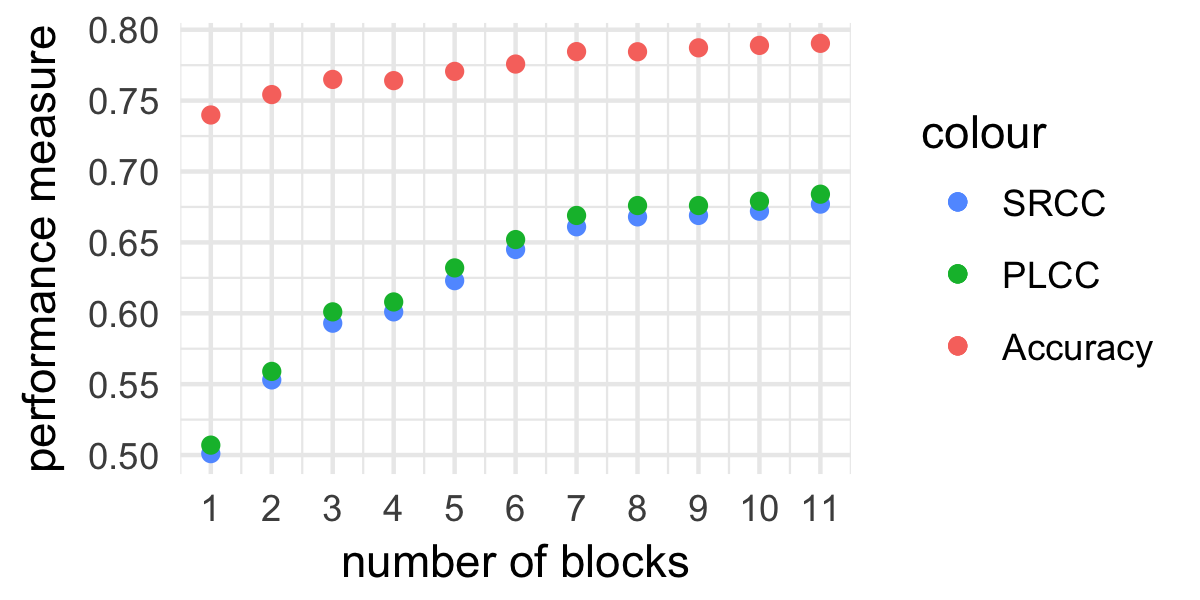}
    \caption{Performance of learning from pre-trained features (Single-3FC model) when increasing the number of inception blocks that are being considered, from the last to the first block (Inception-v3).}
    \vspace{-10pt}
    \label{fig:performance-of-blocks}
\end{figure}

In some of the previous works \cite{karayev2013recognizing,hii_multigap_2017} only the last few levels of the network have been used, 2 and 5 respectively.  However, in Table \ref{tab:feature-architectures}, we have also shown that Single-1FC performs worse when using only the last 5 levels of Inception-v3 compared to all 11. To further investigate this dependency, we evaluate the performance of the Single-3FC model when increasing the number of levels considered, from last (content) to first (low level features) in the network, see Fig.~\ref{fig:performance-of-blocks}. The last 8 layers have a stronger effect on the training performance, however, the first 3 still bring some benefit.

\subsection{Arbitrary input image shape}
Moving on to features extracted from the original images, the results become much better. In Table \ref{tab:features-originals} we can see that the wider MLSP features perform the best (Pool-3FC). Narrow features, extracted from InceptionResNet-v2 based architecture, which is deeper and has a better performance at classification than Inception-v3, perform about as well as the spatially pooled features from Inception-v3. This may also be partially due to the increased granularity that more kernels offer: there are 16,928 kernels extracted from 43 levels in InceptionResNet-v2, compared to 10,048 kernels from 11 levels in Inception-v3.

\begin{table}[h]
\centering
\resizebox{0.9\linewidth}{!}{
\begin{tabular}{@{}lccc@{}}
\toprule
Architecture              & SRCC  & PLCC  & Accuracy \\ \midrule
Single-3FC (-aug)         & 0.722 & 0.726 & 80.31\%  \\
Single-3FC                & 0.729 & 0.733 & 80.65\%  \\
Pool-3FC                  & 0.745 & 0.748 & 81.43\%  \\
Single-3FC (*)(-aug)      & 0.740 & 0.742 & 80.96\%  \\
Single-3FC (*)            & 0.743 & 0.745 & 81.37\%  \\
Pool-3FC (*)(-aug)        & 0.752 & 0.755 & 81.61\%  \\ 
Pool-3FC (*)     & \textbf{0.756} & \textbf{0.757} & \textbf{81.72\%}  \\ \bottomrule
\end{tabular} 
}
\caption{Training on features from the AVA original images, extracted from Inception-v3 an InceptionResNet-v2. The latter is marked with (*). The performance without crop/flip augmentation is lower (-aug = no augmentation). Training Pool-3FC on the wide MLSP features gives the best performance of all methods.}
\vspace{-10pt}
\label{tab:features-originals}
\end{table}


\subsection{Comparison to previous methods}
The main goal of all existing AQA methods trained on AVA is to improve binary classification accuracy. As we will discuss more in-depth later in Sec. \ref{evaluating-performance}, binary accuracy makes an arbitrary and limiting choice to separate low from high quality images at a score of 5 (of 10). Thus, methods that reach an optimal performance with respect to accuracy may not perform as well for the whole range of scores. This is the case with both existing works that report SRCC as well: for Talebi at al. \cite{talebi_nima_2017} and Kong et al. \cite{kong2016photo} the reported SRCC is much lower relative to the accuracy, see Table \ref{tab:comparison-existing-works}. They report 0.612 SRCC with 0.8151 accuracy (0.75 ratio), and 0.558 SRCC with 0.7733 accuracy (0.72 ratio) respectively. Our feature-based model Single-1FC has an SRCC of 0.644 at a similar accuracy to Kong et al. \cite{kong2016photo} of 0.779 (ratio 0.82), whereas our best method Pool-3FC has an SRCC of 0.756 at a similar accuracy to Talebi at al. \cite{talebi_nima_2017} of 0.8161 (ratio 0.92). The higher ratios show that our feature based methods are generalizing better to the entire range of scores.

The reported results for the NIMA architecture (V2) introduced by Talebi et al. \cite{talebi_nima_2017} are using Inception-v2 as the base architecture. We re-trained their model with Inception-v3 to have a fairer comparison with our work (Talebi et al. \cite{talebi_nima_2017} V3 in Table \ref{tab:comparison-existing-works}). The re-trained V3 model has a lower accuracy, however its performance on the correlation metrics increases. The best Earth Mover's Distance (EMD) loss we obtained was 0.07, compared to the reported 0.05 in the original work. This may be one reason for the difference in performance, however it is possible that the Inception-v3 architecture is more suited for correlation metrics.

Some existing works such as \cite{hii_multigap_2017,ma_alamp_2017} use additional information about each image. We only compare methods that use purely information derived from image content. For instance, \cite{hii_multigap_2017} use user comments to enrich the available information for each image. While this latter source is clearly incompatible with image content, Ma et al. \cite{ma_alamp_2017} indirectly use information about the actual resolution of each image by including it in attribute graphs. Thus \cite{ma_alamp_2017} reports a higher accuracy of 82.5\% when using extra information, while their performance based on image content alone is 81.70\%.

We confirm that the resolution of images is indeed informative on our metrics. For instance, the SRCC between the number of pixels of each image (width $\times$ height) and the MOS is 0.188. This is not entirely unexpected, as higher resolution images of the same quality could lead to a better quality of experience, e.g., due to a wider field of view. We also went ahead and trained a small fully-connected 4-layer network ($11 \times 9 \times 7 \times 2$ neurons) with a cross entropy loss, to predict high (MOS $>5$) and low quality images based on two inputs: the width and height of each image, normalized by dividing with the maximum value for each. Due to the imbalance of the test set the baseline for classification is 71.08\%, i.e., assign all images to the high quality class. Our resolution based network slightly improves this performance, reaching an accuracy of 72.40\%.

\begin{table}[h]
\centering
\resizebox{0.9\linewidth}{!}{
\begin{tabular}{@{}llll@{}}
\toprule
Model                                 & SRCC                      & PLCC                      & Accuracy        \\ \midrule
Murray et al. \cite{murray_ava_2012}  & -                         & -                         & 66.70\%  \\
Kao et al. \cite{kao2015visual}               & -                         & -                         & 71.42\%  \\
Lu et al. \cite{lu_deep_2015}                  & -                         & -                         & 74.46\%  \\
Lu et al. \cite{lu2015rating}                    & -                         & -                         & 75.42\%  \\
Hii et al. \cite{hii_multigap_2017}         &                           &                           & 75.76\%  \\
Mai et al. \cite{mai2016composition}   & -                         & -                          & 77.40\%  \\
Kong et al. \cite{kong2016photo}         & 0.558                  & -                          & 77.33\%  \\
Talebi et al. \cite{talebi_nima_2017} V2   & 0.612              & 0.636                   & 81.51\%   \\
Talebi et al. \cite{talebi_nima_2017} V3  & 0.639*             & 0.645*                  &  72.30\%*   \\
Ma et al. \cite{ma_alamp_2017}           & -                          & -                          & 81.70\%   \\ \midrule
Ours (Pool-3FC)                               &   \textbf{0.756}  			    & \textbf{0.757}			     & \textbf{81.72\%}  \\ \bottomrule
\end{tabular}
}
\caption{Performance comparison for existing methods. All numbers are reproduced as reported in the respective works, except for those marked with (*), which we have retrained. We report the performance of the best methods introduced in their respective works which use image content information exclusively.}
\vspace{-10pt}
\label{tab:comparison-existing-works}
\end{table}



%
%
\section{Discussion}

\subsection{Performance evaluation}
\label{evaluating-performance}
The predominant performance measure used in all methods trained on the AVA data set is the binary classification accuracy, as originally defined by Murray et al. \cite{murray_ava_2012}. While easy to compute, this measure is not the most representative for the entire data set. An important use case for an aesthetics assessment model is its ability to establish ranking relationships between the quality of images. This can help to automatically guide enhancement \cite{talebi_nima_2017} or sort collections of images based on aesthetics. We argue that correlation performance metrics such as Spearman Rank-order Correlation Coefficient (SRCC) and Pearson Linear Correlation Coefficient (PLCC) are better suited for the purpose, compared to binary classification accuracy:

1. High quality images in the binary classification as defined in the original work \cite{murray_ava_2012} and all later works are those that have a MOS greater than 5. Considering that the average MOS in AVA is 5.5, this choice is entirely arbitrary. The resulting accuracy  varies greatly with different choices of the binary split. Optimizing classification performance for any particular threshold is thus not representative for the entire range of scores.

2. Random test sets, including the official one used in most previous works, have an unbalanced number of high quality ($\approx$70\%) and low quality ($\approx$30\%) instances. To the best of our knowledge, the binary classification accuracy measure as reported in the related works is not accounting for imbalanced classes. This implies that the minimum accuracy of any method should be around 70\% , since this is the accuracy of a naive classifier that assigns all test images to the high quality class.


We are aware of only two works~\cite{talebi_nima_2017,kong2016photo} that deal with AQA and report correlation-based performance metrics.
Nonetheless, these metrics have been widely applied for Image Quality Assessment (IQA), and do not have the downsides of the binary accuracy measure. They are representative of the entire range of quality scores, do not make arbitrary choices, and for SRCC account for the ranking errors of the predicted and ground-truth scores. Our approach is optimized to achieve the highest performance on correlation metrics. Nonetheless, we report binary accuracy, achieving comparable results with the state of the art.

\subsection{Failure cases}
\begin{figure*}[h]
    \centering
    \includegraphics[width=1\textwidth]{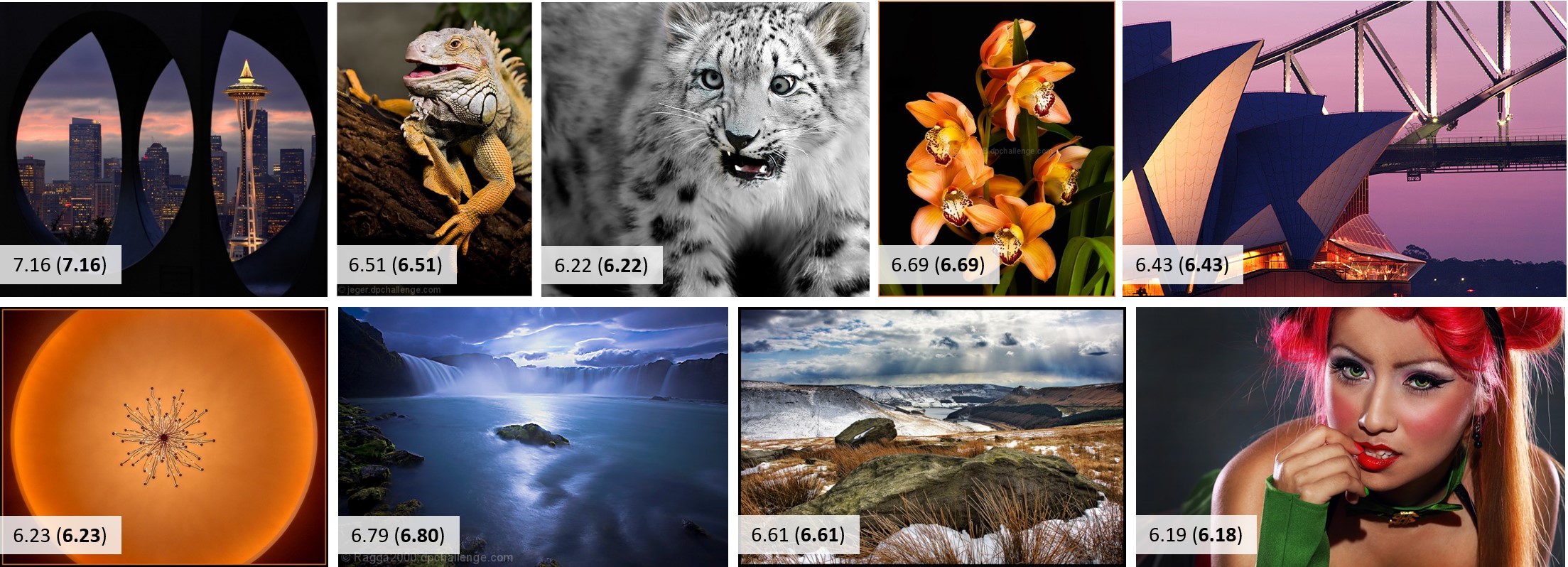}
    \caption{High quality images from the test set, for which our best model's assessment errors are lowest. Our predicted score is the number on the left in each image while the ground-truth MOS is shown in brackets.}
    \label{fig:example-images-low-err}
\end{figure*}

\begin{figure*}[h]
    \centering
    \includegraphics[width=1\textwidth]{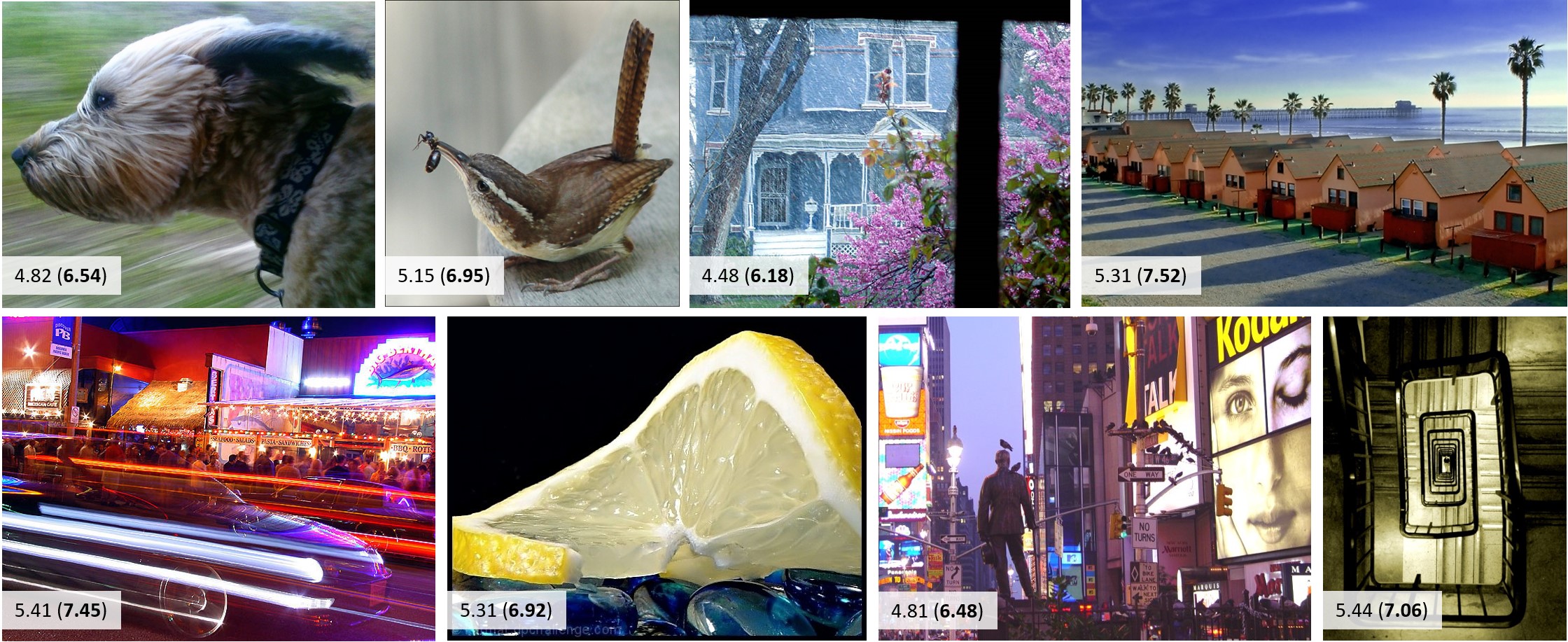}
    \caption{High quality images from the test set, for which our best model's assessment errors are some of the highest.  Our predicted score is the number on the left in each image while the ground-truth MOS is shown in brackets.}
    \label{fig:example-images-high-err}
\end{figure*}

In Figures~\ref{fig:example-images-low-err} and \ref{fig:example-images-high-err} we show several examples of images for which their predicted MOS values are close to the ground-truth (Fig.~\ref{fig:example-images-low-err}) and some that have large absolute errors, see Fig.~\ref{fig:example-images-high-err}. Generally, we observe that images that have a low prediction error have both a high technical quality and show an interesting subject.
In comparison, images that exhibit large prediction errors tend to have some obvious technical faults, however the subject matter is still interesting, thus the high user ratings. It appears our model has an easier time learning to differentiate images based on technical quality aspects, thus over-emphasizing their importance in some unusual cases.

We selected the images as follows. Images in the test set were sorted in decreasing order of their ground-truth MOS. The top 5,000 (out 19,928) were further sorted by the absolute error between the ground-truth MOS and the predicted one. The lowest and highest 40 images were reviewed. The images shown in figures~\ref{fig:example-images-low-err} and \ref{fig:example-images-high-err} have been selected as representative samples. Among low aesthetic quality images, w.r.t. the ground-truth MOS, we could not find a clear pattern that differentiates low and high error cases.


%
%

\section{Conclusions}

Aesthetics quality assessment (AQA) is a challenging and useful perceptual problem. Our proposed approach substantially outperforms the state of the art w.r.t.\ correlation evaluation metrics, and matches it for a less discriminating, but popular metric: classification accuracy. We argued for the use of correlation metrics, which are well established in the broader field of image quality assessment.

Our proposed approach is general, but show-cased on AQA. We introduced a transfer learning strategy that uses features extracted from pretrained ImageNet CNNs (MLSP features, a new type of perceptual features) to surpass the performance of end-to-end training approaches. The power of perceptual features has been suggested before by \cite{gao_blind_2018} and \cite{zhang_unreasonable_2018} in other domains and approaches. We show that our proposed features perform the best for AQA, while enabling much faster training and requiring less GPU memory.


For some problems, the speed gains and low resource usage of employing pre-trained networks can make it possible to learn from larger collections of images, at very high resolutions, and iterate quickly on limited GPU memory. The memory limitation is offset by storing large sets of features on disk. Moreover, using spatially wide and multi-level features leads to further performance gains, suggesting that we are yet to reach the full potential of pre-trained features when applied to perceptual assessment.

While our approach matches the binary classification accuracy of existing works (81.7\%), it is substantially better over the entire score range: our SRCC is 0.756 compared to the existing reported 0.612. The largest performance improvement for our method is achieved by using information extracted from the original resolution images, without rescaling or cropping. This is likely both due to the masking effect of rescaling on technical quality aspects, e.g., noise, blur, sharpness, etc. and the changes in the aesthetics of the composition when cropping small parts of an image.

MLSP features from InceptionResNet-v2 do consistently better than those from Inception-v3. Both of these were trained on a subset of ImageNet (1000 classes). This suggest that deeper architectures, and maybe those trained on larger data sets such as the entire ImageNet, might provide further performance gains when using pre-trained features. 


\section*{Acknowledgment}
Funded by the Deutsche Forschungsgemeinschaft (DFG, German Research Foundation)
in the SFB Transregio 161 "Quantitative Methods for Visual Computing" (Projects A05 and B05),
and the ERC Starting Grant "Light Field Imaging and Analysis" (336978-LIA).

\vfill
\break
{\small
\bibliographystyle{ieee}
\bibliography{egbib}
}

\end{document}

%% file: fig_overview.tex
\begin{figure}[t!]
    \includegraphics[width=\linewidth]{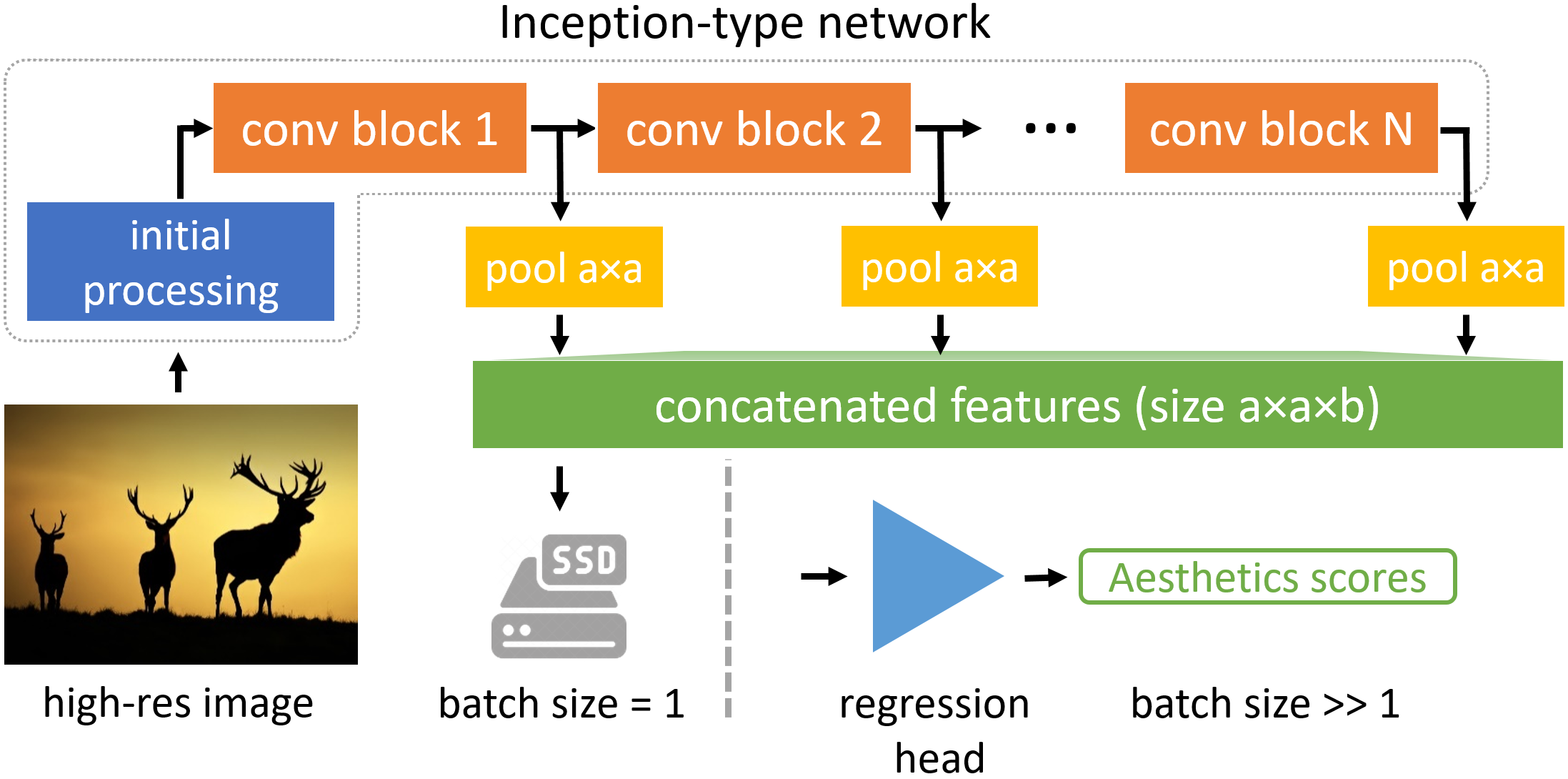}
    \vspace{-5mm}
    \caption{Training pipeline of our framework. We extract and store features from variable sized high resolution images with Inception-type networks, and introduce a new type of features: Multi-level Spatially Pooled activation blocks (MLSP).
    Training aesthetic assessment from features is then done separately, enabling much larger batch sizes. This approach bypasses the resolution bottleneck, and very efficiently trains on the original images in the AVA data set (up to $800\times800$ pixels).}
    \vspace{-3mm}
    \label{fig:overview}
\end{figure}

%% file: fig_3fc_head_arch.tex
\begin{figure}[t!]
    \includegraphics[width=\linewidth]{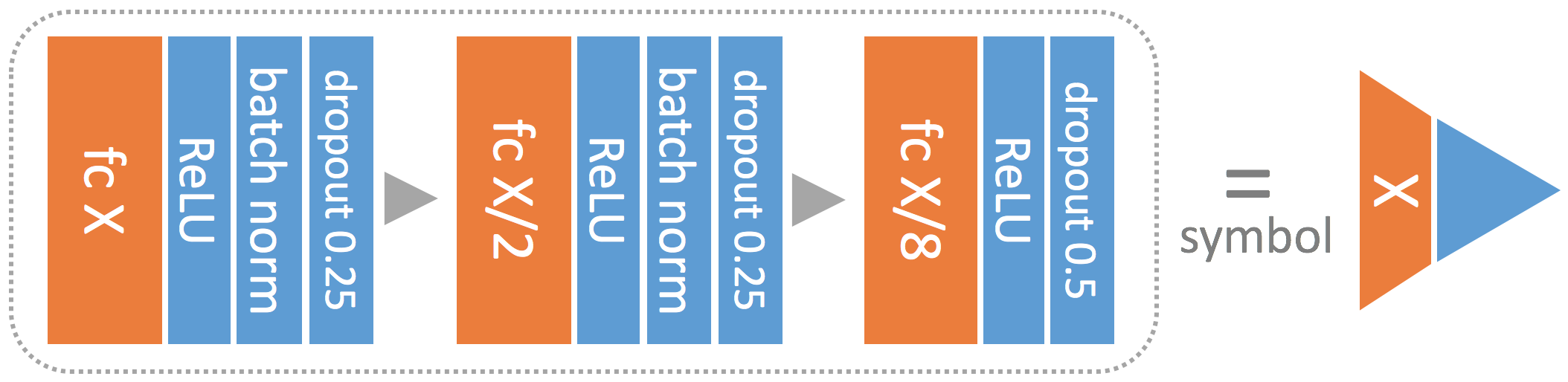}
    \vspace{-5mm}
    \caption{Three layer fully-connected component (3FC) that is used in multiple of our proposed models. The size~$X$ of the first layer depends on the individual model it is used in, ranging from 256 to 2048 neurons.}
    \label{fig:3fc_head_arch}
    \vspace{-3mm}
\end{figure}

%% file: fig_narrow_head_arch.tex
\begin{figure}[b!]
    \centering
    \includegraphics[width=0.75\linewidth]{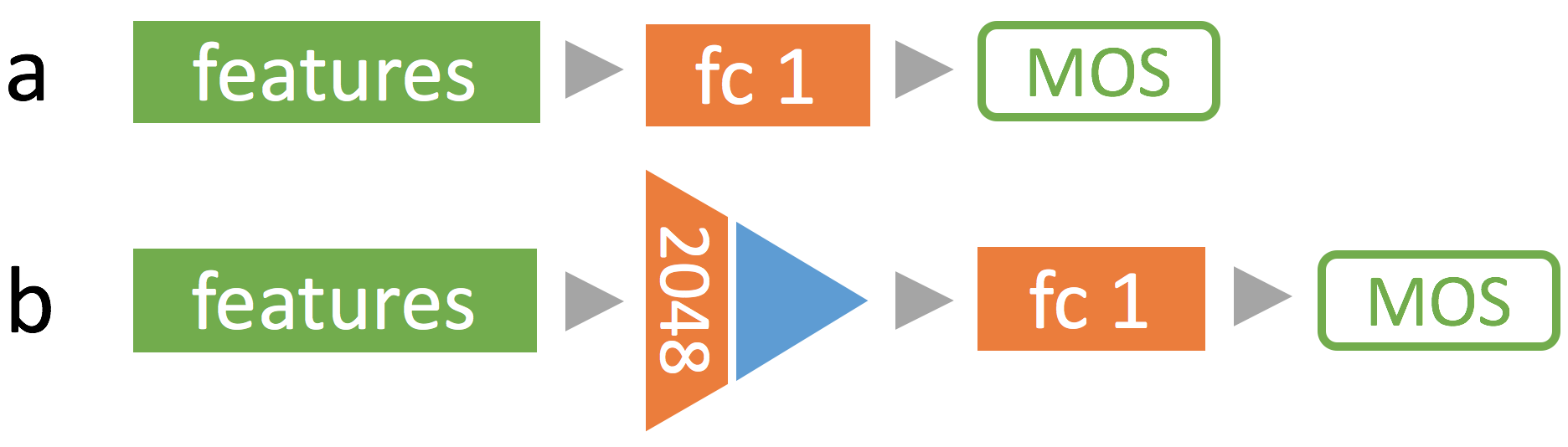}
    \caption{\textbf{a)} \textbf{Single-1FC} architecture, \textbf{b)} \textbf{Single-3FC} with $X=2048$. Complete feature learning architectures used to train on narrow MLSP features ($1\times1\times b$). For each architecture we predict the mean opinion scores (MOS).}
    \label{fig:narrow_head_arch}
\end{figure}

%% file: fig_forest_arch.tex
\begin{figure}[t!]
    \centering
    \includegraphics[width=\linewidth]{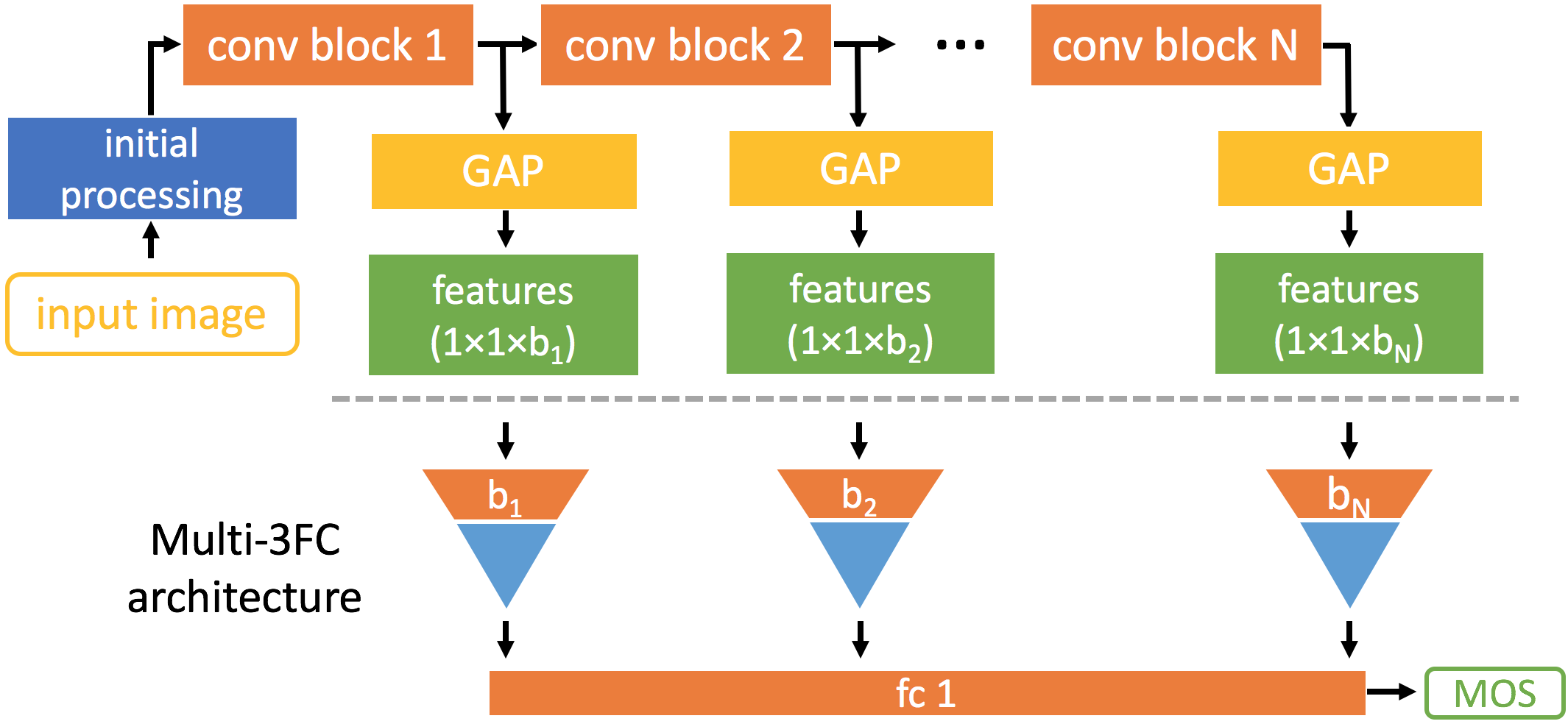}
    \caption{\textbf{Multi-3FC}: alternative architecture for learning from narrow MLSE features. Features extracted from different levels of the network are used separately. The joint prediction is their weighted sum. For each 3FC head (orange-blue triangle), $X=b_i, i=1..N$.}
    \label{fig:forest_arch}
\end{figure}

%% file: fig_spatial_head_architecture.tex
\begin{figure}[h]
    \centering
    \includegraphics[width=0.9\linewidth]{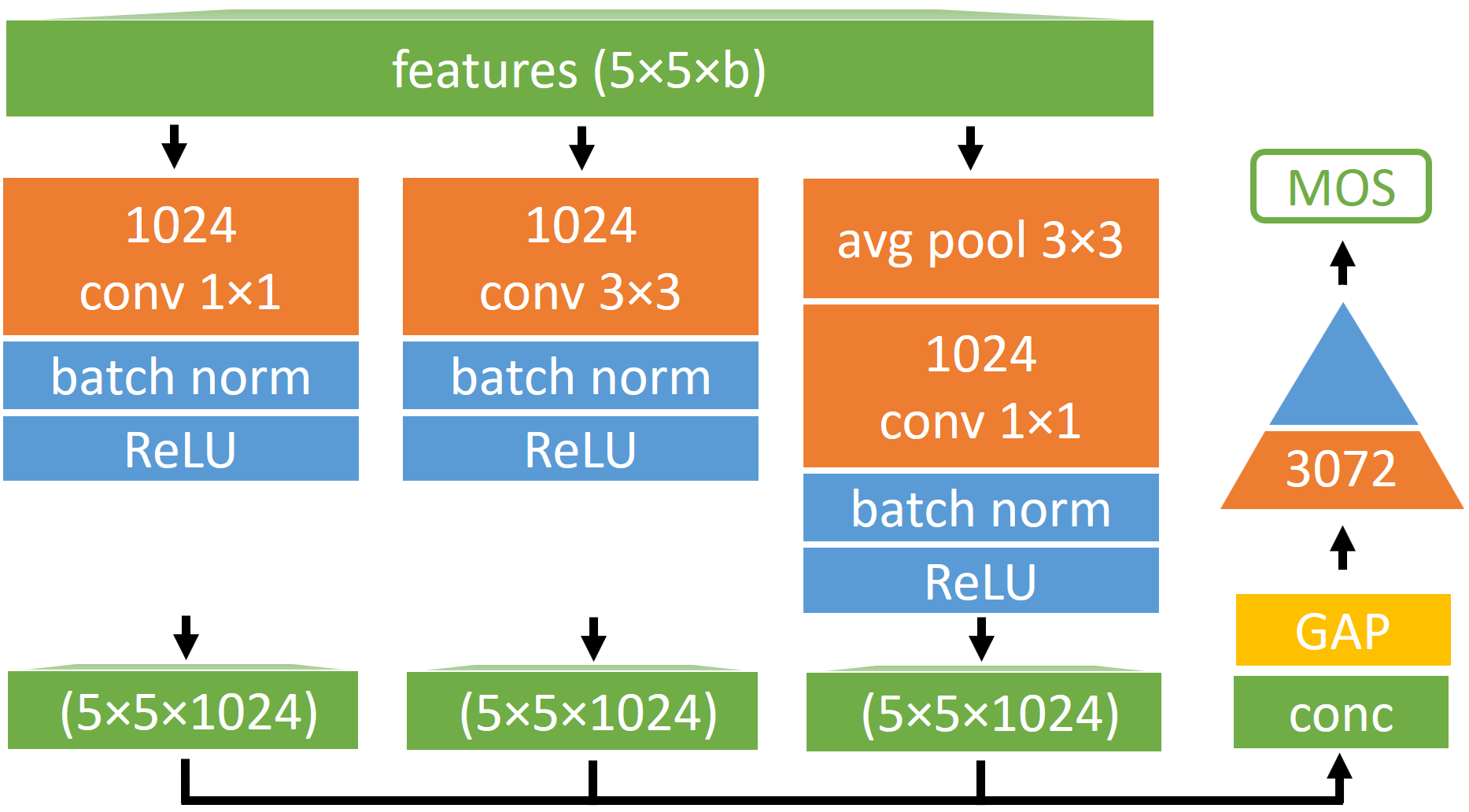}
    \caption{\textbf{Pool-3FC}: Architecture for wide MLSP features. It first reduces the dimension of the features using a modified Inception module (without the $5\times5$ convolution), pools the concatenated features and adds a 3FC head at the end with $X=3072$. The sizes are $b=10,048$ for Inception-v3 and $b=16,928$ for InceptionResNet-v2 features. Operations use padding to keep the spatial dimensions unchanged.}
    \label{fig:spatial_head_architecture}
    \vspace{-3mm}
\end{figure}